\documentclass[11pt]{article}

\usepackage[T1]{fontenc}
\usepackage[utf8]{inputenc}
\usepackage[margin=1in]{geometry}
\usepackage{microtype}
\usepackage{authblk}

\usepackage{amsmath,amssymb,amsfonts,amsthm,mathtools}

\usepackage{graphicx}
\graphicspath{{images/}}
\usepackage{booktabs}
\usepackage{multirow}
\usepackage{array}
\usepackage{tabularx}
\usepackage{longtable}
\usepackage{rotating}
\usepackage{float}

\usepackage{xcolor}
\usepackage{enumitem}

\usepackage[numbers,sort&compress]{natbib}
\usepackage{url}
\usepackage[hidelinks]{hyperref}

\theoremstyle{definition}

\theoremstyle{remark}

\usepackage{algorithm}
\usepackage{algpseudocode}
\usepackage{appendix}

\providecommand{\keywords}[1]{%
  \par\vspace{0.5em}\noindent\textbf{Keywords: }#1\par
}

\title{An Integrated Deep Learning and Statistical Framework for Whole-Network Gene--Environment Association with Leaf Vascular Architecture}

\author[1]{Geran Zhao}
\author[1]{Yangsheng Wang}
\author[2]{Xiaotian Dai}
\author[1]{Guifang Fu\thanks{Corresponding author: gfu@binghamton.edu}}

\affil[1]{%
Department of Mathematics and Statistics, Binghamton University, Binghamton, New York 13902, USA}

\affil[2]{%
Department of Mathematics, Illinois State University, Normal, Illinois 61790, USA}

\date{}
\begin{document}

\maketitle
\abstract{Leaf veins exhibit remarkable diversity in architecture and patterning, yet existing gene--environment association studies have primarily quantified leaf venation using a small collection of low-dimensional summary traits, thereby discarding most of the structural information contained in the original images. We propose an integrated deep learning and statistical framework that, to the best of our knowledge, is the first to investigate gene--environment associations by quantifying leaf vascular architecture as a whole-network. The proposed framework achieves four methodological advances. First, it represents the complete leaf vascular architecture as a whole-network image phenotype, preserving both fine-scale local details and global topological structures. Second, it fine-tunes the deep learning-based Edge Detection with Transformers (EDTER) model to accurately extract whole-network leaf vascular architecture from RGB images by jointly learning local and global contextual features. Compared with traditional image-processing techniques, EDTER reduces the need for specialized image preparation while maintaining robust performance on low-quality images with weak vein contrast. Third, it constructs a new annotated leaf image database by integrating edge maps generated by DiffusionEdge with the Berkeley Segmentation Database (BSDS500), providing a bridge between existing benchmark datasets and user-generated image collections for domain adaptation and specialized edge-detection tasks. Fourth, it applies Semiparametric Sparse Canonical Correlation Analysis (SSCCA) to perform variable selection and model associations between repeatedly measured high-dimensional Bivariate image responses and high-dimensional predictors while simultaneously accommodating sparse, zero-inflated data represented by edge maps through a truncated latent Gaussian copula model. Two simulation studies demonstrate the performance of the proposed framework under increasing levels of complexity. Application to a real \emph{Populus} dataset identifies three significant gene--geography interactions associated with leaf vascular architecture, providing new biological insights and establishing a broadly applicable methodological framework for high-dimensional complex image phenotypes.
}
\keywords{Edge Detection, Leaf Vascular Architecture, Deep Learning, Semiparametric Sparse Canonical Correlation Analysis, Genotype--Phenotype Associations}
\maketitle

\section{Introduction}

Leaf veins exhibit exceptional diversity in architecture and patterning, representing one of the most striking examples of vascular network variation in nature \citep{sack2013leaf}. Veins are typically organized hierarchically into first-order (midvein), second-order (lateral veins), third-order, and additional subtle minor veins \citep{sack2013leaf}. This diversity is manifested in leaf function, hydraulic transport, mechanical support, and evolutionary adaptation. Specifically, leaf veins serve as the primary pathways for water transport, thereby influencing hydraulic function, stomatal behavior, maximum photosynthetic rate, drought tolerance, and hydraulic safety margins \citep{brodribb2010viewing}. Moreover, leaf veins contribute to the mechanical toughness of the lamina by serving as the structural barriers that resist fracture \citep{lucas1991fracture}. Leaf venation is also linked to broader ecological adaptations, including herbivory resistance and variation in leaf lifespan \citep{nardini2022hard}. Existing studies have demonstrated that genes play a critical role in regulating leaf vascular architecture, which is highly heritable and provides a unique opportunity to investigate natural variation and uncover genotype--phenotype relationships \citep{rishmawi2017quantitative, narawatthana2023multi}. 
 
Understanding how genes inlfuence leaf vascular architecture requires not only precise quantification of leaf venation but also accurate downstream statistical modeling, both of which involve high-dimensional phenotypic and genetic data. Leaf vein extraction is particularly challenging because vein structures often appear as fine, subtle, and low-contrast internal edges that are difficult to detect accurately. Existing studies have quantified leaf venation using four broad categories of traits \citep{rishmawi2017quantitative, narawatthana2023multi}: structural, topological, geometric, and mechanical traits. Structural traits, such as vein density (VD), midvein area (MVA), vein number (VN), vein diameter, and hierarchical vein orders, primarily determine water transport efficiency \citep{sack2013leaf}. Topological traits, including areole density, loopiness, and connectivity, influence hydraulic redundancy and damage resilience \citep{blonder2011venation}. Geometric traits, such as vein length, orientation, and branching angles, describe the spatial organization of leaf venation, and influence leaf folding patterns and hydraulic pathways \citep{price2011leaf}. Mechanical and anatomical traits, including bundle sheath dimensions, sclerenchyma investment, and leaf toughness, contribute to leaf durability, herbivory resistance, and leaf lifespan \citep{onoda2011global}. However, as low-dimensional summaries extracted from the original images, they provided only a partial representation of the vascular architecture, thereby discarding most of the rich structural information contained in the full venascular architecture.

Several well-established tools have been developed to extract leaf vein traits from images, including phenoVein, LIMANI, LEAF GUI, and NEFI \citep{price2011leaf, dhondt2012quantitative, buhler2015phenovein, dirnberger2015nefi}. These tools extracted low-dimensional venation traits such as vein length, vein density, branching points, and areole area, etc. They primarily relied on image-processing techniques involving segmentation, skeletonization, and graph reconstruction rather than learning-based approaches. Consequently, their performance was often sensitive to image quality and performed best on images with high vein contrast. Segmentation errors can lead to disconnected veins, false branches, or inaccurate trait extraction. Moreover, some methods required specialized image preparation. For example, LIMANI was developed for chemically cleared leaves, in which pigments and soft tissues were chemically removed to enhance the recognition of the leaf vein traits. 

Few learning-based approaches have been developed for leaf venation analysis. \citet{lagergren2023few} employed a few-shot convolutional neural network (CNN) to segment vein structure from high-resolution RGB leaf images. However, they still subsequently extracted low-dimensional summary traits, such as vein length and vein density, from the segmented vein images using RhizoVision Explorer (RVE), rather than directly modeling the full venation architecture.

After low-dimensional leaf vein traits have been extracted, downstream gene--vein association analyses in the existing literature primarily relied on mixed linear models (MLM), fixed and random model circulating probability unification (FarmCPU), and Bayesian-information and linkage-disequilibrium iteratively nested keyway (BLINK) \citep{yu2006unified, liu2016iterative, huang2019blink}. \citet{rishmawi2017quantitative} employed MLM and Markov models to identify candidate genes associated with VD, areole number, and the number of vein endpoints in \emph{Arabidopsis thaliana}. \citet{narawatthana2023multi} applied MLM, FarmCPU, BLINK, and haplotype analysis to identify genetic associations with the distance between minor veins (IVD) in rice. While these statistical approaches successfully identified gene--vein associations, they primarily focused on low-dimensional traits, and were not directly applicable to high-dimensional image traits containing the whole-network. Furthermore, these approaches were largely restricted by parametric structures or normality assumptions.

In this article, we propose an integrated deep learning and statistical framework that achieves four methodological advances driven by gene--vein association analysis. First, we quantify the complete leaf vascular architecture as a whole-network phenotype rather than a small collection of low-dimensional summary traits. Unlike conventional approaches, the proposed framework preserves the complete structural information of the vascular network, including both fine-scale local details and global network topology, and creates a new framework for investigating gene--environment associations with the leaf vascular whole-network phenotype.

Second, we fine-tune the Edge Detection with Transformers (EDTER), a deep learning-based approach that overcomes several limitations of traditional handcrafted image-processing techniques by reducing the need for specialized image preparation and maintaining robust performance on low-quality images with weak vein contrast. EDTER takes RGB images of the leaf lamina as input and outputs a pixel-level edge map of the same spatial dimension by jointly learning global and local contextual features \citep{pu2022edter}. CNN-based methods, such as Holistically-Nested Edge Detection (HED) and Richer Convolutional Features (RCF), improved upon traditional image-processing approaches by learning multi-scale hierarchical feature representations \citep{xie2015holistically, liu2017richer}. However, CNNs primarily relied on local receptive fields and therefore had limited ability to capture long-range spatial dependencies. This limitation is particularly relevant for leaf vein extraction, where vein networks may extend across large spatial regions and exhibit complex global connectivity patterns. Transformer-based edge detection methods address this challenge through self-attention mechanisms that directly model long-range dependencies and global contextual information.

Third, the widely used edge-detection benchmark, the Berkeley Segmentation Database (BSDS500) \citep{arbelaez2010contour}, contains relatively few leaf images, which may limit the ability of the pretrained EDTER models to accurately identify fine leaf veins. To address this limitation, we construct a new leaf image database and generate high-quality annotated edge maps using DiffusionEdge, a state-of-the-art edge-detection framework based on Diffusion Probabilistic Models \citep{ye2024diffusionedge}. Although the edge maps generated by DiffusionEdge are not manually annotated ground-truth labels, they provide an efficient and scalable source of pseudo-ground-truth supervision. This strategy offers a practical alternative to labor-intensive manual annotation while enabling rapid construction of domain-specific training datasets. More importantly, it establishes a bridge between existing benchmark datasets and user-generated image collections, facilitating domain adaptation and fine-tuning for specialized edge-detection tasks.

Fourth, instead of relying on parametric statistical models for low-dimensional trait associations, we apply Semiparametric Sparse Canonical Correlation Analysis (SSCCA) to model the relationships between repeatedly measured high-dimensional Bivariate images and high-dimensional predictors. The edge maps are high-dimensional images in which the majority of pixels correspond to background and only a small fraction represent vein structures. After normalizing pixel intensities to the interval $[0,1]$, most background pixels take values of zero or near zero, resulting in highly sparse and zero-inflated image responses. SSCCA incorporates a truncated latent Gaussian copula model to estimate the correlation structure of high-dimensional zero-inflated data \citep{yoon2020sparse}. 

We conduct two simulation studies with increasing levels of complexity to evaluate the performance of the proposed framework. We further apply the proposed framework to a real \emph{Populus} dataset consisting of 100 leaves, each photographed from both the front and back sides, resulting in 200 RGB leaf images, together with 118 genetic- and geography-related predictor variables. Because the front and back sides of the same leaf share the same genetic and geographic information, the corresponding edge maps are treated as repeatedly measured Bivariate images for each subject. To the best of our knowledge, this is the first study to employ deep learning techniques to represent leaf vascular architecture as a whole-network image phenotype for gene--environment association analysis, and the identified associations provide new biological insights into genotype--phenotype relationships. The proposed framework provides a general methodology for the analysis of broader network-structured and image-based data.

\section{Methods}\label{sec2}

\subsection{The Structure of EDTER}\label{subsec1}
\begin{figure}
    \centering
    \includegraphics[width=1\linewidth]{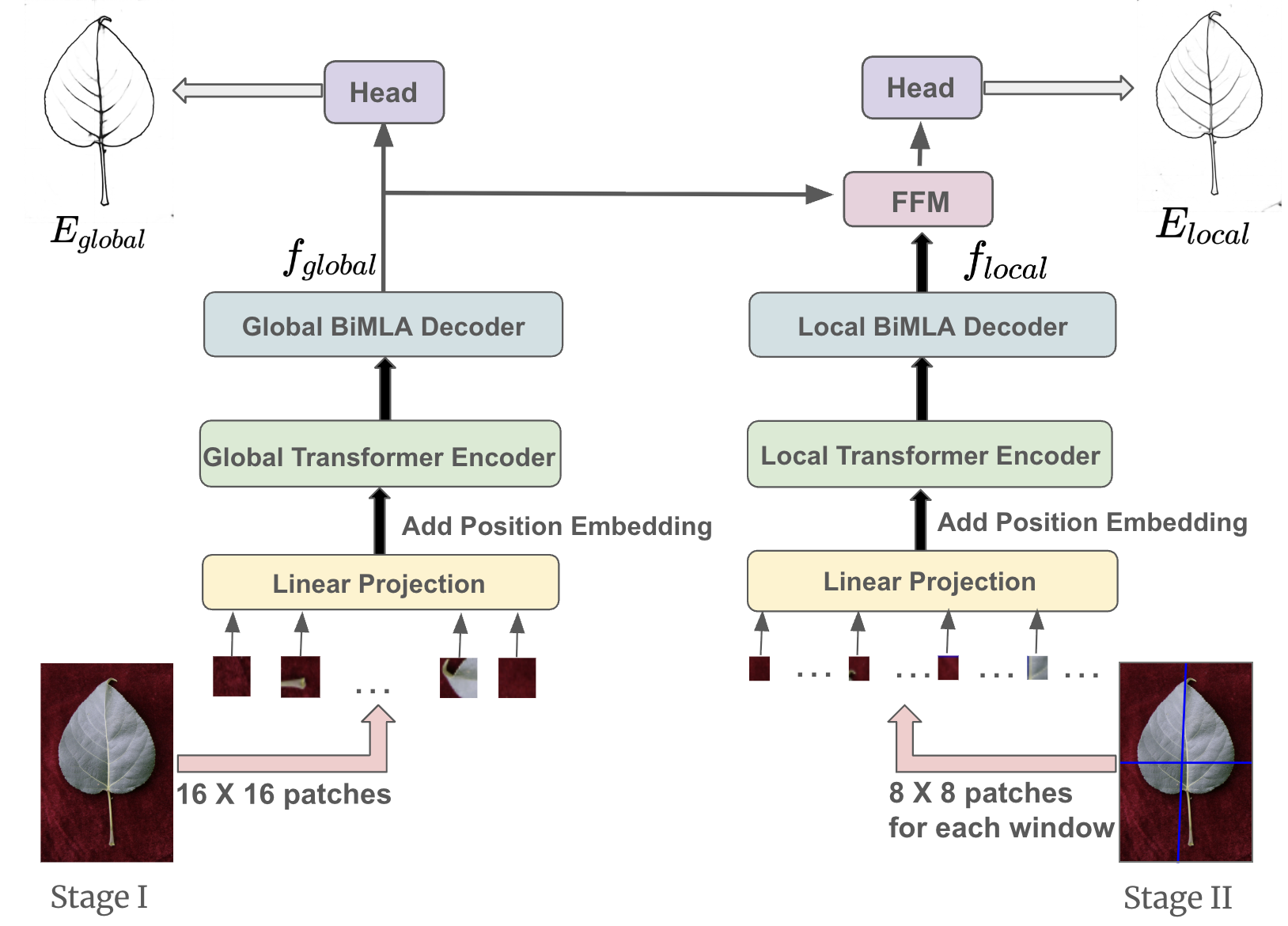}
    \caption{A schematic overview of EDTER. $E_{local}$ denotes the final edge map produced by EDTER through the fusion of global and local feature representations.}
    \label{fig:EDTER}
\end{figure}

Edge detection methods are commonly designed to identify object boundaries and internal edges in images. EDTER is a Transformer-based edge detector that combines global contextual information with local fine-grained cues through a two-stage architecture. Specifically, Stage I is designed to learn global contextual information by employing a global transformer encoder and a global bidirectional multi-level aggregation (BiMLA) decoder to generate high-resolution feature representations. Stage II focuses on learning short-range local cues and applies a local transformer encoder and a local BiMLA module to produce pixel-level feature maps. The outputs from both stages are integrated through a Feature Fusion Module (FFM) to predict the final edge maps \citep{pu2022edter}. See Figure \ref{fig:EDTER} for a schematic overview of EDTER. 

\subsubsection{Transformer Encoder} \label{subsubsec1}
In the EDTER framework, the transformer encoder adopts the Vision Transformer (ViT) architecture, which represents an image as a sequence of embedded image patches and processes them through multiple transformer blocks \citep{dosovitskiy2020image}. For Stage I, the input RGB image \(I\in\mathbb{R}^{H\times W\times3}\) is partitioned into a sequence of non-overlapping \(16\times16\) image patches, where \(H\) and \(W\) denote the height and width of the image. For Stage II, the same input image is first divided into four non-overlapping windows, each of size \(\frac{H}{2}\times\frac{W}{2}\). Each window is then partitioned into a sequence of non-overlapping \(8\times8\) image patches. The resulting patches are flattened and projected into a latent embedding space through a learnable linear projection. Positional embeddings are subsequently added to preserve spatial information.

The transformer encoder consists of a sequence of transformer blocks, each composed of Layer Normalizations (LayerNorm), a Multi-head Self-Attention (MSA) module, residual connections, and a Multi-Layer Perceptron (MLP). Let \(M\) denote the number of attention heads and \(L\) denote the number of transformer blocks. After patch embedding and positional embedding, the initial token embedding is denoted by \(Z_0\). Let \(Z_\ell\) denote the output representation of the \(\ell\)th transformer block. The transformer blocks are connected recursively through
$Z_{\ell}=\operatorname{Block}_{\ell}(Z_{\ell-1}),~\ell=1,\ldots,L,$ where \(Z_{\ell}\in\mathbb{R}^{N\times J}\), \(N\) is the number of image patches (tokens), and \(J\) is the embedding dimension of each token. Here, \(\operatorname{Block}_{\ell}(\cdot)\) denotes the complete \(\ell\)th transformer block, consisting of LayerNorm, an MSA, residual connections, and an MLP.

Within the \(\ell\)th transformer block, Layer Normalization is first applied to the input, $\widetilde{Z}_{\ell-1} =\operatorname{LayerNorm}(Z_{\ell-1}).$ The normalized token embeddings serve as the common input to all \(M\) attention heads. For the \(m\)th attention head (\(m=1,\ldots,M\)), the query, key, and value matrices are computed as 
\[
\begin{aligned}
Q_{\ell}^{(m)}
&=
\widetilde{Z}_{\ell-1}W_{Q,\ell}^{(m)},\\
K_{\ell}^{(m)}
&=
\widetilde{Z}_{\ell-1}W_{K,\ell}^{(m)},\\
V_{\ell}^{(m)}
&=
\widetilde{Z}_{\ell-1}W_{V,\ell}^{(m)},
\end{aligned}
\]
where $W_{Q,\ell}^{(m)}, W_{K,\ell}^{(m)},W_{V,\ell}^{(m)} \in \mathbb{R}^{J\times d}$ are learnable projection matrices for each of the attention head in the \(\ell\)th transformer block, and \(d\) denotes the dimension of each attention head.

The output of the \(m\)th attention head in the \(\ell\)th transformer block is computed as
\[
O_{\ell}^{(m)}
=
\operatorname{softmax}
\!\left(
\frac{
Q_{\ell}^{(m)}
(K_{\ell}^{(m)})^\top
}
{\sqrt d}
\right)
V_{\ell}^{(m)},
\]
which are concatenated across all \(M\) attention heads and projected back to the embedding dimension,
\[
O_{\ell}
=
\left[
O_{\ell}^{(1)},
O_{\ell}^{(2)},
\ldots,
O_{\ell}^{(M)}
\right]
W_{O,\ell},
\]
where \(W_{O,\ell}\in\mathbb{R}^{Md\times J}\) is the projection matrix of the \(\ell\)th MSA.

Note that $O_{\ell}$ is the output of a MSA module and it is subsequently processed through residual connections, LayerNorm, and an MLP to produce the output representation \(Z_{\ell}\) of the \(\ell\)th transformer block. In this study, we set \(M=16\). The global transformer encoder consists of \(L=24\) transformer blocks, producing the sequence of output representations $\{Z_1,\ldots,Z_{24}\}$, whereas the local transformer encoder consists of \(L=12\) transformer blocks, producing $\{T_1,\ldots,T_{12}\}.$

\subsubsection{Bi-directional Multi-Level Aggregation Decoder} \label{subsubsec2}
\begin{figure}
    \centering
    \includegraphics[width=\linewidth]{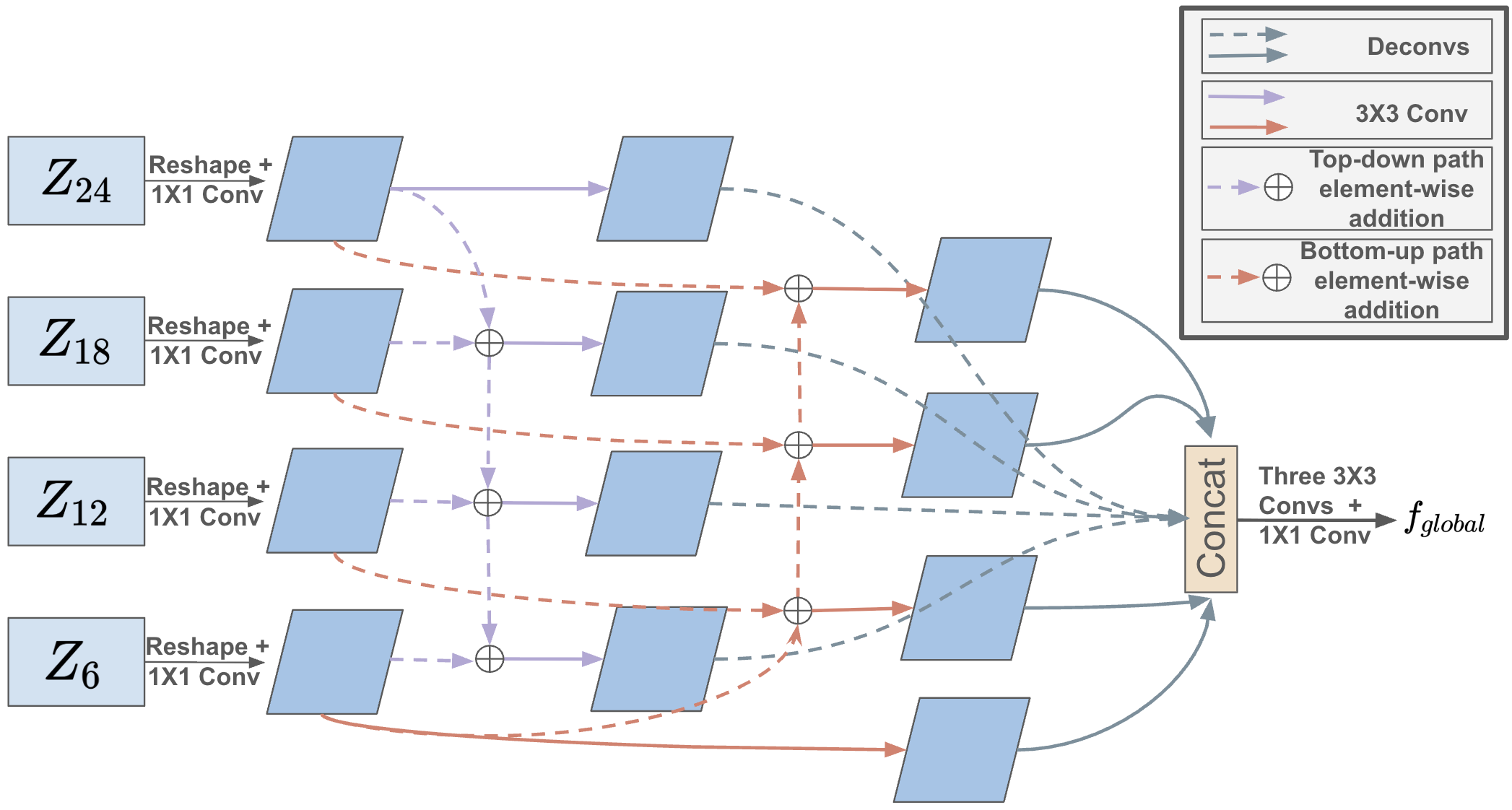}
    \caption{A schematic overview of the Bi-directional Multi-Level Aggregation Decoder.}
    \label{fig:BiMLA}
\end{figure}

The global transformer encoder produces 24 transformer blocks, which are evenly divided into four groups. The last block from each group, namely \(Z_6\), \(Z_{12}\), \(Z_{18}\), and \(Z_{24}\), is selected as the input to the BiMLA decoder. Each output is reshaped from an \(N\times J\) matrix into a three-dimensional feature map of size \(H/16\times W/16\times J\), restoring the spatial arrangement of the image patches. A subsequent \(1\times1\) convolution aligns the feature channels before feature aggregation. As illustrated in Figure~\ref{fig:BiMLA}, the top-down path propagates high-level semantic information through successive feature fusion operations, whereas the bottom-up path propagates low-level spatial details while preserving fine edge information. Together, these complementary paths effectively aggregate multi-scale contextual information.

To recover spatial resolution, deconvolution layers with \(4\times4\) and \(16\times16\) kernels are employed to upsample feature maps from different scales. Each deconvolution layer is followed by Batch Normalization (BN) and a ReLU activation function. The upsampled feature maps are subsequently concatenated and refined by three consecutive \(3\times3\) convolution layers followed by a final \(1\times1\) convolution layer, producing the global edge feature map \(f_{global}\).

Figure~\ref{fig:BiMLA} illustrates a schematic overview of the global BiMLA decoder. The local BiMLA decoder follows the same overall design, except that the \(3\times3\) convolution layers are replaced by \(1\times1\) convolution layers to reduce the introduction of artificial edges while preserving local edge details. The local BiMLA takes \(T_3\), \(T_6\), \(T_9\), and \(T_{12}\) as its inputs and outputs edge feature map $f_{local}$.

\subsubsection{Feature Fusion Module} \label{subsubsec3}
\begin{figure}
    \centering
    \includegraphics[width=1\linewidth]{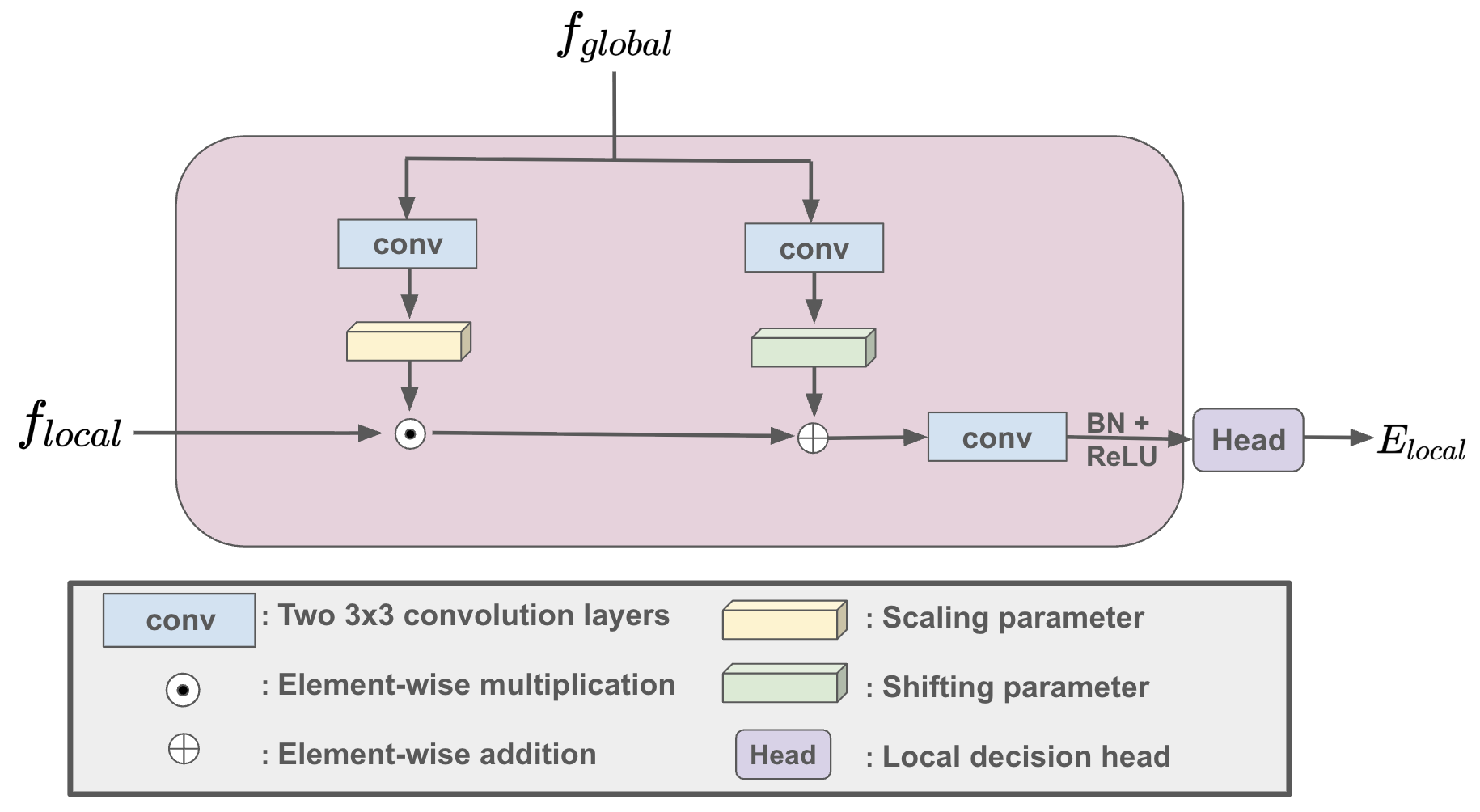}
    \caption{A schematic overview of the Feature Fusion Module. }
    \label{fig:ffm}
\end{figure}

To integrate the complementary information extracted from the global and local transformer encoders, EDTER employs a Feature Fusion Module, illustrated in Figure~\ref{fig:ffm}. The FFM is built upon a Spatial Feature Transform (SFT) layer \citep{wang2018recovering}, followed by convolutional, Batch Normalization, and ReLU operations. The global feature map $f_{global}$ is first processed by two parallel convolution branches to generate two sets of spatially adaptive modulation parameters. These parameters provide contextual guidance for refining the local feature representation and are applied to the local feature map $f_{local}$ through element-wise multiplication followed by element-wise addition. The resulting fused feature map is then processed by an additional $3\times3$ convolutional layer, followed by BN and ReLU, before being passed to the local decision head to predict the refined edge map.
\subsection{Fine-tuning with Enlarged New Leaf Database}\label{subsec2}
\begin{figure*}
\centering
\includegraphics[width=0.8\linewidth]{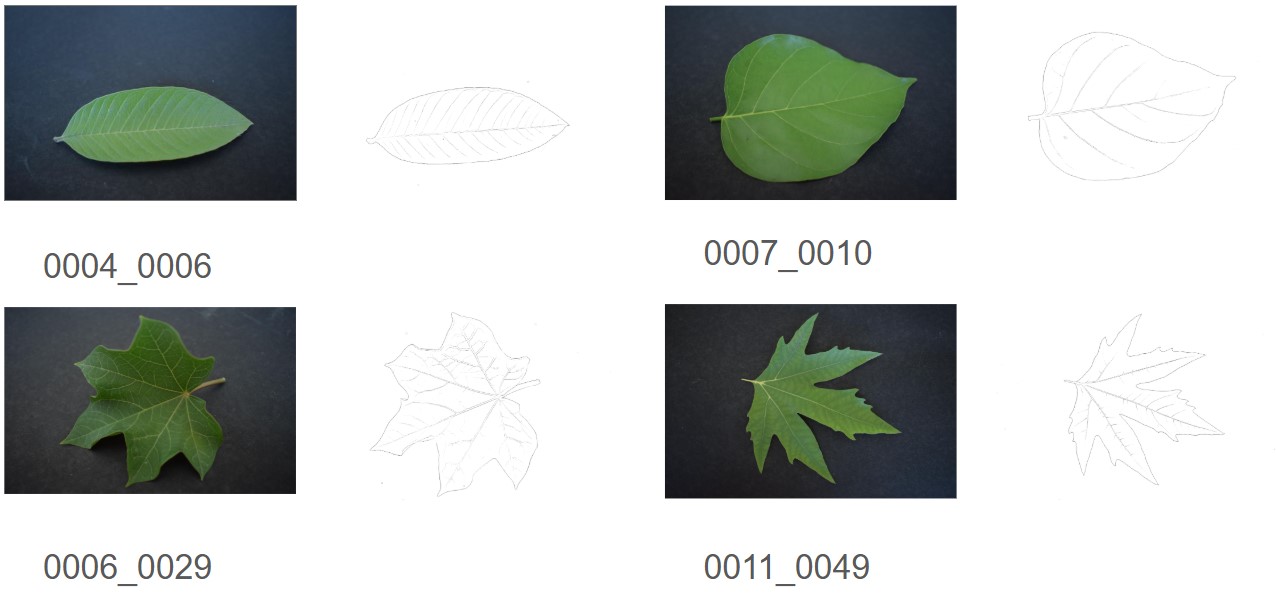}
\caption{Four randomly selected examples from the newly constructed leaf image dataset, together with the corresponding RGB images and edge maps annotated by DiffusionEdge.}
\label{fig:leafdata}
\end{figure*}

We construct a new leaf image database consisting of 324 selected healthy leaves from the Mendeley Leaf Image Database \citep{siddharth2019database}. However, these images do not contain manually annotated edge maps, which are required as ground truth labels for fine-tuning EDTER. Unlike deterministic edge-detection methods, DiffusionEdge formulates edge detection as a conditional image generation task, progressively refining edge maps from Gaussian noise under the guidance of the original RGB image. Figure~\ref{fig:leafdata} presents four examples from the new leaf database together with the corresponding edge maps annotated by DiffusionEdge, which exhibit sharp and thin vein structures that visually resemble the manually annotated edge maps provided in BSDS500. 

To improve the model's adaptation to leaf venation images, we combine this dataset with BSDS500 and utilize the annotated edge maps as supervisory labels for fine-tuning EDTER \citep{arbelaez2010contour}. We initialize the model with pre-trained weights provided by EDTER and further train it for 20,000 more iterations for stage I and 40,000 more iterations for stage II. The fine-tuned EDTER model is subsequently applied to the leaf images in the real dataset to generate venation edge maps for downstream association analysis.

\begin{figure}
\centering
\includegraphics[width=0.8\linewidth]{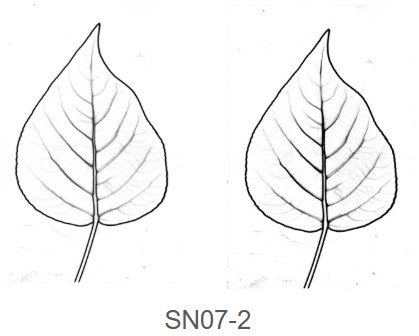}
\caption{Comparison of the EDTER model before and after fine-tuning. The left panel illustrates the output of the pre-trained EDTER model, whereas the right panel shows the output of the fine-tuned EDTER model.}
\label{fig:fine-tune}
\end{figure}

Figure~\ref{fig:fine-tune} compares the edge maps generated by the pretrained and fine-tuned EDTER models. The fine-tuned EDTER model (right panel) preserves substantially more minor veins and fine-scale branching structures, resulting in a more complete representation of the whole-network. In addition, vein segments of the right panel appear more continuous and better connected than the left panel, reducing fragmentation in the extracted venation patterns. These improvements are particularly evident in regions with dense tertiary and quaternary venation, where many faint vein segments become clearly noticeable after fine-tuning. 

\subsection{Semiparametric Sparse Canonical Correlation Analysis }\label{subsec3}
Let \(\mathbf{A}^{(1)}, \mathbf{A}^{(2)} \in \mathbb{R}^{H\times W}\) denote the edge maps extracted by EDTER from the front and back sides of the same leaf, respectively. The original edge maps produced by EDTER assign pixel values close to 0 to vein structures and values close to 1 to background regions. To improve interpretability of the downstream statistical analysis, we invert the pixel intensities so that vein pixels take values close to 1, whereas background pixels take values close to 0. Throughout this article, \(\mathbf{A}^{(1)}\) and \(\mathbf{A}^{(2)}\) refer to the inverted edge maps.

Let \(\mathbf{Y}^{(1)}, \mathbf{Y}^{(2)} \in \mathbb{R}^{p_2/2}\) denote the vectorized representations of \(\mathbf{A}^{(1)}\) and \(\mathbf{A}^{(2)}\), respectively, obtained by concatenating all image pixels into vectors, where \(p_2/2=H\times W\). We define the Bivariate image response as
\[
\mathbf{Y}
=
\begin{bmatrix}
\mathbf{Y}^{(1)}\\
\mathbf{Y}^{(2)}
\end{bmatrix}
\in\mathbb{R}^{p_2}.
\]
Let \(\mathbf{X}=(X_{1},X_{2},\ldots,X_{p_1})\in\mathbb{R}^{p_1}\) denote the high-dimensional predictor vector. 

Classical canonical correlation analysis (CCA) is a widely used statistical method for modeling the association between two sets of vectors \citep{hotelling1992relations}. CCA seeks the canonical weight vectors \(\boldsymbol{\omega}_x\) and \(\boldsymbol{\omega}_y\) that maximize the correlation between the linear combinations \(\boldsymbol{\omega}_x^\top \mathbf{X}\) and \(\boldsymbol{\omega}_y^\top \mathbf{Y}\) by solving
\begin{equation}
\label{cca1}
\max_{\boldsymbol{\omega}_x,\boldsymbol{\omega}_y}
\ \boldsymbol{\omega}_x^\top \boldsymbol{\Sigma}_{xy}\boldsymbol{\omega}_y
\quad
\text{s.t.}
\quad
\boldsymbol{\omega}_x^\top\boldsymbol{\Sigma}_x\boldsymbol{\omega}_x=1,\qquad
\boldsymbol{\omega}_y^\top\boldsymbol{\Sigma}_y\boldsymbol{\omega}_y=1,
\end{equation}
where \(\boldsymbol{\Sigma}_x=\operatorname{cov}(\mathbf{X})\), \(\boldsymbol{\Sigma}_y=\operatorname{cov}(\mathbf{Y})\), and \(\boldsymbol{\Sigma}_{xy}=\operatorname{cov}(\mathbf{X},\mathbf{Y})\). Classical CCA is not directly applicable to high-dimensional settings because the sample covariance matrices become singular or ill-conditioned, resulting in unstable or non-unique solutions. To overcome this limitation, numerous sparse CCA methods have been proposed by incorporating an \(\ell_1\) penalty into the canonical weight vectors \citep{witten2009penalized}. However, these methods are primarily developed for continuous data and do not explicitly account for the highly sparse and zero-inflated characteristics of edge-map images. Therefore, we employ the Semiparametric Sparse CCA, which is specifically designed to accommodate high-dimensional zero-inflated sparse data \citep{yoon2020sparse}. Rather than relying on the conventional sample covariance matrix, SSCCA estimates the correlation structure under a truncated latent Gaussian copula model.

\textbf{Definition \citep{yoon2020sparse} (Truncated Latent Gaussian Copula model).} The image (or Bivariate image) response $\mathbf Y$ is assumed to follow a truncated latent Gaussian copula model if there exists a latent random vector 
\[
\mathbf{U}=(U_1,\ldots,U_{p_2})^\top
\sim
\mathrm{NPN}(\mathbf{0},\boldsymbol{\Sigma},f),
\]
such that
$
Y_j=I(U_j>C_j)U_j,~j=1,\ldots,p_2,$ where \(I(\cdot)\) denotes the indicator function, \(\mathbf{C}=(C_1,\ldots,C_{p_2})^\top\) is a vector of truncation constants, and \(\boldsymbol{\Sigma}\) is the latent correlation matrix. Here, \(\mathrm{NPN}\) denotes the nonparanormal distribution, and \(f=\{f_j,~j=1,\ldots,p_2\}\) represents a collection of monotonically increasing transformation functions. Consequently, $\mathbf{Y}\sim\mathrm{TLNPN}(\mathbf{0},\boldsymbol{\Sigma},f,\mathbf{C}),$ where \(\mathrm{TLNPN}\) denotes the truncated latent nonparanormal distribution.

To estimate the latent correlation matrix between the zero-inflated response $\mathbf{Y}$ and the continuous predictor $\mathbf{X}$, SSCCA employs a bridge function that relates Kendall's $\tau$ to the latent correlation coefficient. Specifically, the latent correlation between $Y_j$ and $X_k$ is given by $\Sigma_{jk}=F^{-1}(\tau_{jk}),$ where $F(\cdot)$ denotes the bridge function, whose explicit form for this mixed data type is given below \citep{yoon2020sparse},
\begin{equation*}
    F(\Sigma_{jk}; \Delta_j) = -2\Phi_2 \left( -\Delta_j, 0; \frac{1}{\sqrt{2}} \right) + 4\Phi_3(-\Delta_j, 0, 0; \Sigma'_3),
\end{equation*}
and
\begin{equation*}
    \Sigma'_3 = \begin{pmatrix}
    1 & 1/\sqrt{2} & \Sigma_{jk}/\sqrt{2} \\
    1/\sqrt{2} & 1 & \Sigma_{jk} \\
    \Sigma_{jk}/\sqrt{2} & \Sigma_{jk} & 1
    \end{pmatrix}.
\end{equation*}
Here, $\Phi_d(\cdots,\Sigma'_d)$ denotes the cdf of the standard $d$-variate normal distribution with correlation matrix $\Sigma'_d$. The latent threshold $\Delta_j=f_j(C_j)$ is estimated by $\widehat{\Delta}_j=\Phi^{-1}(1-\widehat{\pi}_j)$, where $\widehat{\pi}_j$ is the proportion of non-zero observations in $Y_j$.

Let $n$ denote the sample size and let $\{(X_{ik},Y_{ij})\}_{i=1}^n$ be independent and identically distributed observations for $Y_j$ and $X_k$. The sample Kendall's $\tau$ is defined as
\[
\widehat{\tau}_{jk}
=
\frac{2}{n(n-1)}
\sum_{1\le i<i'\le n}
\operatorname{sign}(Y_{ij}-Y_{i'j})
\operatorname{sign}(X_{ik}-X_{i'k}).
\]
Its population counterpart satisfies $\tau_{jk}=\mathbb{E}(\widehat{\tau}_{jk})=F(\Sigma_{jk},\Delta_j).$ Accordingly, SSCCA estimates the population latent correlation $\Sigma_{jk}$ by $\widehat r_{jk}$, where
$\widehat r_{jk}=F^{-1}(\widehat\tau_{jk})$
is obtained by replacing the population Kendall's $\tau_{jk}$ with its sample counterpart $\widehat{\tau}_{jk}$ through the inverse bridge function. The resulting rank-based correlation estimator is
$\widehat{\mathbf R}=(\widehat r_{jk}).$

To ensure positive definiteness, \(\widehat{\mathbf R}\) is regularized as
\[
\widetilde{\mathbf R}
=
(1-\mu)\widehat{\mathbf R}^{\,\mathrm{PSD}}
+
\mu\mathbf I,
\]
where $\mathbf I$ is the identity matrix, \(\mu\) is a small positive threshold (set to 0.01 in this article), and
\(\widehat{\mathbf R}^{\,\mathrm{PSD}}\) denotes the projection of
\(\widehat{\mathbf R}\) onto the cone of positive semidefinite matrices
\citep{fan2017high,yoon2020sparse}.

Finally, the SSCCA optimization problem is formulated as
\begin{equation}
\label{eq2}
\begin{split}
\min_{\boldsymbol{\omega}_x,\boldsymbol{\omega}_y}\;
&
\left\{
-\boldsymbol{\omega}_x^\top
\widetilde{\mathbf{R}}_{xy}
\boldsymbol{\omega}_y
+\lambda_1\|\boldsymbol{\omega}_x\|_1
+\lambda_2\|\boldsymbol{\omega}_y\|_1
\right\}
\\
\text{s.t.}\quad
&
\boldsymbol{\omega}_x^\top
\widetilde{\mathbf{R}}_x
\boldsymbol{\omega}_x
\le1,\qquad
\boldsymbol{\omega}_y^\top
\widetilde{\mathbf{R}}_y
\boldsymbol{\omega}_y
\le1.
\end{split}
\end{equation}

This constrained optimization problem is equivalent to solving
\[
\widetilde{\boldsymbol{\omega}}_x
=
\arg\min_{\boldsymbol{\omega}_x}
\left\{
\frac12
\boldsymbol{\omega}_x^\top
\widetilde{\mathbf{R}}_x
\boldsymbol{\omega}_x
-
\boldsymbol{\omega}_x^\top
\widetilde{\mathbf{R}}_{xy}
\boldsymbol{\omega}_y
+\lambda_1
\|\boldsymbol{\omega}_x\|_1
\right\}.
\]

The solution is then normalized by setting
\[
\widehat{\boldsymbol{\omega}}_x=
\begin{cases}
\mathbf{0},
&
\widetilde{\boldsymbol{\omega}}_x=\mathbf{0},
\\[2mm]
\displaystyle
\frac{\widetilde{\boldsymbol{\omega}}_x}
{\left(
\widetilde{\boldsymbol{\omega}}_x^\top
\widetilde{\mathbf{R}}_x
\widetilde{\boldsymbol{\omega}}_x
\right)^{1/2}},
&
\widetilde{\boldsymbol{\omega}}_x\neq\mathbf{0}.
\end{cases}
\]

Another attractive property of SSCCA is its ability to perform variable selection through an $\ell_1$ (LASSO) penalty, which shrinks the coefficients of non-informative $\mathbf{X}$ toward zero while retaining only the predictors most strongly associated with $\mathbf{Y}$. The tuning parameters \(\lambda_1\) and \(\lambda_2\) are selected using the Bayesian Information Criterion (BIC) proposed by \citet{yoon2020sparse}, defined as
\begin{equation}
\mathrm{BIC}
=
g(\widetilde{\boldsymbol{\omega}}_x)
+
df_{\widetilde{\boldsymbol{\omega}}_x}
\frac{\log n}{n},
\end{equation}
where $
g(\widetilde{\boldsymbol{\omega}}_x)
=
\widetilde{\boldsymbol{\omega}}_x^\top
\widetilde{\mathbf{R}}_x
\widetilde{\boldsymbol{\omega}}_x
-
2\widetilde{\boldsymbol{\omega}}_x^\top
\widetilde{\mathbf{R}}_{xy}
\boldsymbol{\omega}_y
+
\boldsymbol{\omega}_y^\top
\widetilde{\mathbf{R}}_y
\boldsymbol{\omega}_y,
$
and \(df_{\widetilde{\boldsymbol{\omega}}_x}\) denotes the number of nonzero entries in \(\widetilde{\boldsymbol{\omega}}_x\). By the symmetry of the objective function, $\boldsymbol{\omega}_y$ can be estimated analogously. The SSCCA model is implemented using the \textbf{mixedCCA} \textsf{R} package.

\section{Simulation}\label{sec3}
We design two simulation scenarios of increasing complexity to evaluate the accuracy of SSCCA to identify the true predictors associated with the simulated leaf vein image data. The following four evaluation metrics are adopted \citep{li2012feature}.
\begin{itemize}
    \item $\mathcal{S}$ is defined as the average number of predictors selected by the model across 100 replications. 
   \item $\mathcal{P}_i$ is defined as the proportion of each individual true predictor being selected by the model across 100 replications.
   \item $\mathcal{P}_s$ is defined as the proportion of all true predictors being simultaneously selected by the model across 100 replications.
    \item $\mathcal{FSR}$ is defined as the false selection rate that represents the proportion of noise predictors that are incorrectly selected by the model across 100 replications.
\end{itemize}
\subsection{Simulation 1}\label{subsec3_2}
We set \(n=200\) and \(p_1=50\). The predictor vectors $\mathbf{X}$ are generated from $\mathbf{X} \sim \mathcal{N}(\mathbf{0},\boldsymbol{\Sigma}_x)$, where \(\boldsymbol{\Sigma}_x=(\sigma_{k_1k_2})\) is defined by $\sigma_{k_1k_2}=\rho^{|k_1-k_2|},~\rho=0.8.$ To mimic the categorical nature of genetic markers, each component of \(\mathbf{X}\) is discretized into three categorical levels. Let \(X_k\) denote the \(k\)th predictor variable. We define the set of true predictors as $S^*=\{X_6,X_{13},X_{21},X_{34},X_{41}\},$ with corresponding coefficients $\beta_6=1,~ \beta_{13}=-2,~
\beta_{21}=-1,~ \beta_{34}=1,~ \beta_{41}=0.5.$ All remaining coefficients are set to zero, that is, $\beta_k=0,~k\notin S^*.$

To generate the leaf vein image response data, we select one edge map extracted by EDTER from the real dataset as a template image, whose pixel values are denoted by \(T_{s,t}\), \(s=1,\ldots,H\) and \(t=1,\ldots,W\). For subject \(i\), the simulated image \(\mathbf{A}_i=(A_{i,s,t})\) is generated by perturbing the template image with a predictor-dependent signal and random noise according to
\[
A_{i,s,t}
=
T_{s,t}
+
\gamma\Big(\sum_{k=1}^{50}\beta_k X_{ik}\Big)
+
\epsilon_{i,s,t},
\]
where \(\epsilon_{i,s,t}\) are i.i.d. \(\mathcal{N}(0,1)\), \(\gamma=10\). Finally, the pixel intensities of $\mathbf{A}_i$ are truncated to the range $A_{i,s,t}\in [0,255]$ to satisfy the valid intensity range of grayscale images.

\begin{figure}
    \centering
    \includegraphics[width=0.8\linewidth]{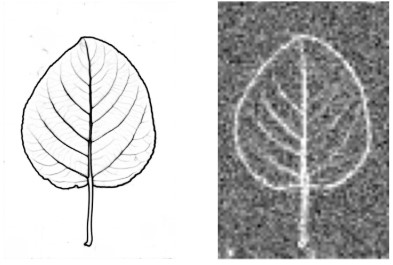}
    \caption{Illustration of Simulation 1. The left panel shows the edge map of the reference leaf, whereas the right panel presents a randomly selected simulated leaf generated from the reference edge map.}
    \label{fig:sim1}
\end{figure}

Figure~\ref{fig:sim1} presents an example of an image generated under Simulation~1. In this setting, the predictor-dependent signal is added uniformly across all pixels, affecting both   background regions and vein structures. As shown in Figure~\ref{fig:sim1}, the resulting images are heavily contaminated by noise, making the underlying minor veins difficult to distinguish from the background.

\begin{table*}[!t]
\centering
\caption{Results of Simulation 1. $\mathcal{S}$ denotes the average model size, $\mathcal{P}_i$ the individual success rate, $\mathcal{P}_s$ the simultaneous success rate, $\mathcal{FSR}$ the false selection rate, and SD the sample standard deviation computed from 100 simulation replicates. \label{tab1}}
\begin{tabular}{clllllcc}
\hline
$\mathcal{S}$ (SD)                         
& \multicolumn{5}{c}{$\mathcal{P}_i$ (SD)}                                 
& $\mathcal{P}_s$ (SD)                             
& $\mathcal{FSR}$ (SD)           \\ \hline
\multicolumn{1}{c|}{\multirow{2}{*}{5.72 (0.86)}} 
& $X_6$ & $X_{13}$ & $X_{21}$ & $X_{34}$ & \multicolumn{1}{l|}{$X_{41}$} 
& \multicolumn{1}{c|}{\multirow{2}{*}{0.99 (0.1)}} 
& \multirow{2}{*}{0.016 (0.189)} \\ \cline{2-6}
\multicolumn{1}{c|}{}                      
& 1 (0) & 1 (0) & 0.99 (0.1) & 1 (0) & \multicolumn{1}{l|}{1 (0)}    
& \multicolumn{1}{c|}{}                            
&                                \\ \hline
\end{tabular}
\end{table*}

The results of Simulation~1 are summarized in Table~\ref{tab1}. The average model size, \(\mathcal{S}\), selected by SSCCA is 5.72, which is very close to the true number of active predictors, indicating high variable selection accuracy. The individual success rate, \(\mathcal{P}_i\), for each of the five true predictors is remarkably high. In particular, \(X_6\), \(X_{13}\), \(X_{34}\), and \(X_{41}\) are selected 100\% with SD = 0. The simultaneous success rate, \(\mathcal{P}_s\), reaches 99\%, indicating near-perfect discovery of all true signals simultaneously. Meanwhile, the false selection rate, \(\mathcal{FSR}\), remains low at 0.016, demonstrating effective control of false discoveries.


\subsection{Simulation 2}\label{subsec3_3}
To increase the complexity level, Simulation~2 restricts the predictor effects to vein regions only. Since vein pixels typically have higher intensity values than background pixels, we introduce a threshold \(\alpha\) and allow the five true predictors to influence only those pixels with intensities exceeding \(\alpha\) (in this article, we set \(\alpha=80\)). Specifically, the response images are generated according to
\begin{equation}
A_{i,s,t}
=
T_{s,t}
+
\gamma \Big( \sum_{k=1}^{50} \beta_k X_{ik} \Big)
I(T_{s,t}>\alpha)
+
\epsilon_{i,s,t},
\label{sim2}
\end{equation}
where \(I(\cdot)\) denotes the indicator function, \(\epsilon_{i,s,t}\) are i.i.d. \(\mathcal{N}(0,1)\), and the resulting pixel intensities are truncated to the interval \([0,255]\). All remaining parameters and coefficients are set to be identical to those in Simulation~1. 

\begin{figure}
    \centering
    \includegraphics[width=0.8\linewidth]{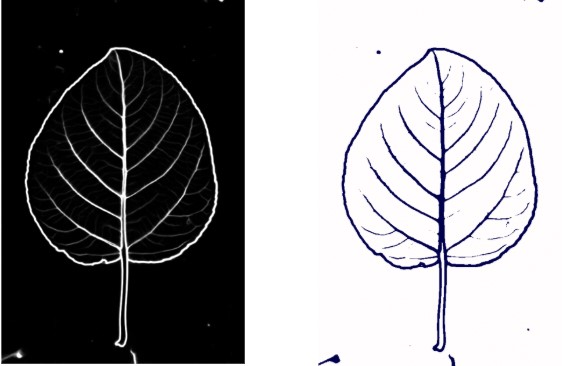}
    \caption{Illustration of Simulation 2. The left panel shows the edge map of the reference leaf, whereas the right panel illustrates the spatial locations and intensities of the simulated predictor signal and random noise added to the reference edge map.}
    \label{fig:sim2}
\end{figure}

The left panel of Figure~\ref{fig:sim2} displays the original template leaf edge map \(T_{s,t}\), whereas the right panel shows the added signal (i.e., the second and third term in equation (\ref{sim2})), which enables us to visualize the regions where the predictor-related signals and random noise are introduced, as well as their relative intensities.

Table~\ref{tab2} summarizes the results of Simulation~2. The average model size, \(\mathcal{S}=5.55\), remains very close to the true number of active predictors, indicating accurate variable selection performance. Compared with Simulation~1, the association signal is restricted to a small subset of pixels located on or near the vein structures, making signal discovery substantially more challenging. But SSCCA still achieves a $91\%$ simultaneous success rate while maintaining a low false selection rate of $\mathcal{FSR}=0.021$.

\begin{table*}[!t]
\centering
\caption{Results of Simulation 2. $\mathcal{S}$ denotes the average model size, $\mathcal{P}_i$ the individual success rate, $\mathcal{P}_s$ the simultaneous success rate, $\mathcal{FSR}$ the false selection rate, and SD the sample standard deviation computed from 100 simulation replicates. \label{tab2}}
\begin{tabular}{clllllcc}
\hline
$\mathcal{S}$ (SD)                         
& \multicolumn{5}{c}{$\mathcal{P}_i$ (SD)}                                 
& $\mathcal{P}_s$ (SD)                             
& $\mathcal{FSR}$ (SD)           \\ \hline
\multicolumn{1}{c|}{\multirow{2}{*}{5.55 (1.01)}} 
& $X_6$ & $X_{13}$ & $X_{21}$ & $X_{34}$ & \multicolumn{1}{l|}{$X_{41}$} 
& \multicolumn{1}{c|}{\multirow{2}{*}{0.91 (0.29)}} 
& \multirow{2}{*}{0.021 (0.03)} \\ \cline{2-6}
\multicolumn{1}{c|}{}                      
& 0.91 (0.29) & 0.91 (0.29) & 0.92 (0.28) & 0.93 (0.26) & \multicolumn{1}{l|}{0.92 (0.27)}    
& \multicolumn{1}{c|}{}                            
&                                \\ \hline
\end{tabular}
\end{table*}

\section{Real Data Analysis}\label{sec4}
The real data analyzed in this study were collected from a population of \emph{Populus szechuanica} var. \emph{tibetica}, a member of the \emph{Tacamahaca} section of the genus \emph{Populus}. Native to the Tibetan Plateau, this species is distributed across Gansu, Shaanxi, Sichuan, Xizang, and Yunnan provinces of China. The sampled trees span a broad geographical range, with latitudes from $28^\circ58.337' \mathrm{N}$ to $29^\circ58.915' \mathrm{N}$, longitudes from $88^\circ52.283' \mathrm{E}$ to $94^\circ47.903' \mathrm{E}$, and elevations ranging from 2540 to 4081 meters \citep{fu2013mapping}. The leaves of \emph{P. szechuanica} var. \emph{tibetica} exhibit substantial variation in both morphology and vascular architecture, making them an ideal system for investigating the genetic and environmental factors regulating leaf vascular network.

\begin{figure*}[!t]
\centering
\includegraphics[width=0.6\linewidth]{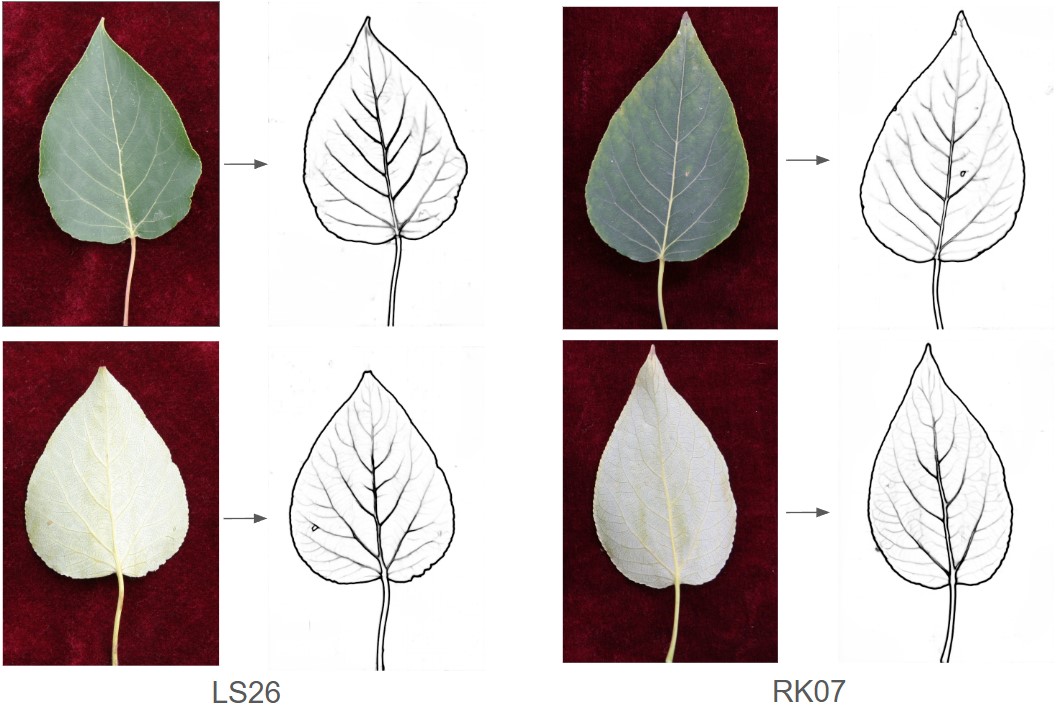}
\caption{Two randomly selected leaves from the real \emph{Populus} dataset. Each column corresponds to one leaf, with the top and bottom panels showing the front and back sides, respectively. For each side, the original RGB image and the corresponding edge map generated by the fine-tuned EDTER model are displayed side by side.}
\label{fig:real3}
\end{figure*}

For each of the \(\emph{Populus}\) trees under study, one leaf was randomly selected and photographed from both the front and back sides. To characterize the vascular whole-network of each leaf, we apply the fine-tuned EDTER model to the RGB images and generate corresponding edge maps. Figure~\ref{fig:real3} presents two representative leaf samples together with their edge maps. The extracted edge maps preserve the whole-network. In some cases, the edge maps generated from the back side capture substantially more vein details than those obtained from the front side, justifying the inclusion of both sides in the analysis.

To reduce computational cost, each Bivariate image is resized from its original dimension of \(900\times600\times2\) to \(60\times40\times2\). The predictor set contains 118 standardized predictors, including 28 categorical genetic markers, 3 continuous geographic variables (latitude, longitude, and elevation), 84 gene--geography interaction terms, and 3 geography--geography interaction terms. We then apply SSCCA to investigate the associations between the Bivariate edge-map response images and the 118 predictor variables.

Among the 31 main effects and 87 interaction terms, SSCCA selects three variables as nonzero canonical coefficients, all of which correspond to gene--geography interaction effects: \emph{GCPM\_1053-1} $\times$ longitude, \emph{GCPM\_1036-1} $\times$ elevation, and \emph{GCPM\_1131} $\times$ latitude. These results suggest that gene--environment interactions play a very important role in influencing the vascular whole-network of \emph{Populus} leaves. 

In addition to identifying new scientific findings, our results also reconfirm some previous studies that were based on low-dimensional vein traits. For example, several leaf venation traits were reported to vary with elevation in woody plants, with trees at higher elevations generally exhibiting lower vein density but greater vein thickness and volume \citep{wang2020plant}. Similarly, \citet{zhu2012pattern} found that minor vein density in natural populations of oriental oak decreased significantly with increasing latitude. A study of three herbaceous species further reported that leaf major vein thickness increased with elevation \citep{liu2020variation}. 
 


\section{Conclusion}\label{sec6}

The proposed framework provides four methodological advances over existing approaches for gene--environment association studies of leaf vascular architecture. By leveraging state-of-the-art deep learning techniques, the proposed framework preserves the complete whole-network of the leaf vascular architecture that cannot be adequately captured by a limited set of summary traits. In addition, the transformer-based EDTER model reduces the dependence on specialized image preparation and demonstrates robust performance in extracting vein structures from images with weak contrast. From a statistical perspective, the SSCCA model enables the joint analysis of repeatedly measured high-dimensional Bivariate image responses and high-dimensional predictors, while simultaneously performing variable selection and accommodating the sparse and zero-inflated nature of edge-map images. 

The limitation of the real dataset analyzed in this study is its relatively small sample size and low marker resolution, which may limit the ability to identify specific gene names. Nevertheless, as a methodology-oriented study, our primary objective is to develop and evaluate a new framework that is readily applicable to larger datasets with high-dimensional image responses and predictors. Beyond leaf venation, the proposed framework can be extended to a broad range of network-structured and image-based phenotypes, including root system architecture, human vascular networks, neuroimaging, and transportation networks. It is particularly well suited for applications where manually annotated ground-truth labels are scarce or unavailable.

Although SSCCA is designed for high-dimensional predictors and responses, its computational efficiency may deteriorate when applied to ultra-high-dimensional image data due to memory limitations, computational burden, and increased processing time. To address this challenge, we reduce the image dimension through resizing prior to the downstream association analysis. We experimentally evaluate several image resolutions, including $180\times120$, $120\times80$, $90\times60$, and $60\times40$, through simulation studies. The results indicate that the $60\times40$ resolution achieves the most favorable balance between computational efficiency and statistical accuracy. Consequently, all analyses presented in this article, including both simulation studies and the real data application, are conducted using images resized to $60\times40$. Compared with alternative dimension-reduction approaches such as principal component analysis (PCA), image resizing preserves the complete image information, facilitating biological interpretation. While pixel level details may be smoothed during resizing, the overall venation architecture remains well preserved, making resizing a practical and interpretable strategy for handling ultrahigh-dimensional image responses.


\section{Competing interests}
The authors declare that they have no competing interests.


\bibliographystyle{plainnat}
\bibliography{reference}

\end{document}